**Detection of small changes in medical and random-dot images comparing self-organizing map performance to human detection**

John Mwangi Wandeto[s, d], Henry Nyongesa[d], Yves Rémond[s] and Birgitta Dresp-Langley[s]

[s]ICube UMR 7357 CNRS University of Strasbourg, 4 rue Blaise Pascal, CS 90032

F-67081 Strasbourg Cedex - FRANCE

[d]Dedan Kimathi University of Technology, Nyeri – Mweiga Road, P.O. Box 657-10100, Nyeri - KENYA

Corresponding author: John Mwangi Wandeto, john.wandeto@etu.unistra.fr

**Abstract**

Radiologists use time-series of medical images to monitor the progression of a patient's conditions. They compare information gleaned from sequences of images to gain insight on progression or remission of the lesions, thus evaluating the progress of a patient's condition or response to therapy. Visual methods of determining differences between one series of images to another can be subjective or fail to detect very small differences. We propose the use of quantization errors obtained from self-organizing maps (SOM) for image content analysis. We tested this technique with MRI images to which we progressively added synthetic lesions. We have used a global approach that considers changes on the entire image as opposed to changes in segmented lesion regions only. We claim that this approach does not suffer from the limitations imposed by segmentation, which may compromise the results. Results show quantization errors increased with the increase in lesions on the images. The results are also consistent with previous studies using alternative approaches. We then compared the detectability ability of our method to that of human novice observers having to detect very small local differences in random-dot images. The quantization errors of the SOM outputs compared with correct positive rates, after subtraction of false positive rates ("guess rates"), increased noticeably and consistently with small increases in local dot size that were not detectable by humans. We conclude that our method detects very small changes in complex images and suggest that it could be implemented to assist human operators in image-based decision making.

**Keywords:** medical images; random-dot images; change detection; SOM analysis; quantization error; human performance

## 1. Introduction

Radiologists have to detect the progression of patients' conditions on the basis of, often hardly detectable, local changes in medical images. The images are captured through various imaging techniques, such as magnetic resonance imaging (MRI), computerized tomography (CT) and positron emission tomography (PET). These images provide the radiologist with visual information about the state or progression of a given condition, and help determine the course of treatment. Traditional methods for handling such images involve direct visual inspection, which is by its nature subjective. Image science has proposed methods for the automated processing of medical images, which involves various different image processing techniques to identify specific diagnostic regions of interest and features, such as lesions. [1], [2] proposed a computational framework to enable comparison of MRI volumes based on gray-scale normalization to determine quantitative tumor growth between successive time intervals. They proposed three tumor growth indices, namely, volume, maximum radius and spherical radius. The approach, however, requires an initial manual segmentation of images, which can be a time-consuming task. [3], first, semi-automatically segmented a tumor in an initial patient scan and then aligned the successive scans using a hierarchical registration scheme to measure growth or shrinkage from the images. This method relies on accurate segmentation and requires manual supervision, in order to detect changes of up to a few voxels in the pathology. [4] describe a procedure aimed for difficult-to-detect brain tumor changes. The approach combines input from a medical expert with a computational technique. In this paper, we propose a new technique based on self-organized mapping that considers the whole medical image, as opposed to an image segment, as region of interest. This excludes manual benchmarking tasks designed to eliminate inclusion of structures with similarity to tumor pathology. The basic principle behind direct image analysis is that there exists an intrinsic relationship between medical images and their clinical measurements, which can be exploited to eliminate intermediate procedures in image analysis. Compared to traditional methods, direct methods have more clinical significance by targeting the final outcome. Thus, direct methods not only reduce high computational costs, but also avoid errors induced by any intermediate operations. Direct methods also serve as a bridge between emerging machine learning algorithms and clinical image measurements. Finally, to show how the output variable called "quantization error" of image analysis by SOM may be exploited as an indicator for the presence of potentially critical local changes in image contents, we compared the quantization errors of SOM outputs from analyses of random-dot images with very small progressive increases in the local size of a single dot to the capacity of human observers to detect these changes.

## 2.0 Materials and Methods

### 2.1 *Self-organizing maps*

A self-organizing map (SOM) is an unsupervised neural network learning technique that does not need target outputs required in error correction supervised learning. SOM, [5] are used to produce a lower-dimension representation of the input space. Thus, for each input vector, so called, competitive learning is carried out to produce a lower-dimension visualization of the input data. SOM are typically applied as feature classifiers of input data. From an initial randomization of a map, input data is iteratively applied to optimize the map into stable regions. Where the node weights match the input vector, that area of the lattice is selectively optimized to more closely resemble the data for the class the input vector is a member of. From an initial distribution of random weights and over multiple iterations the SOM eventually settles into a map of stable zones. Each region of the map becomes a feature class of the input space. Each zone is effectively a feature classifier, and the graphical output is a type of feature map of the input space.

FIGURE 1

The central idea behind the principles and mathematics of SOM is that every input data item shall be matched to the closest fitting region of the map, called the winner (as denoted by $M_c$ in Fig. 1), and such subsets of regions shall be modified for optimal matching of the entire data set, [6]. On the other hand, since the spatial neighborhood around the winner in the map is modified at a time, a degree of local and differential ordering of the map occurs to provide a smoothing action. The local ordering actions will gradually be propagated over the entire SOM. The parameters of the SOM models are variable and are adjusted by learning algorithms such that the maps finally approximate or represent the similarity of the input data. While studies have mainly concentrated on the performance of various SOM on a given dataset, we set to unveil the behavior of various datasets on a single SOM. Given related sets of medical image series and a constant SOM, can we detect a significant trend in the images? Is the trend of any clinical significance?

*2.2 The quantization error in SOM outputs*

The task of finding a suitable subset that describes and represents a larger set of data vectors is called vector quantization (VQ), [7]. VQ aims at reducing the number of sample vectors or at substituting them with representative centroids. The resulting centroids do not necessarily have to be from the set of samples but can also be an approximation of the vectors assigned to them, for example their average. VQ is closely related to clustering, and SOM performs VQ since the sample vectors are mapped to a (smaller) number of prototype vectors, [8]. The prototype vectors are called the best matching units (BMU) in SOM. As a property of SOM, the quantization error (QE) is used to evaluate the quality of SOM. The QE belongs to a type of measures that have been used to benchmark a series of SOMs trained from the same dataset. In our work, we have used QE to do a somewhat opposite measure: to benchmark a series of datasets using SOM trained with the same parameters. In other words, we use the same SOM, same map size, feature size, learning rate and neighborhood radius to analyze series of image datasets with clinical significance, or random-dot images, as shown later herein. The QE is derived after subjecting an image to a self-organizing map algorithm analysis and by calculating the squared distance (usually, the standard Euclidean distance) between an input data, $x$, and its corresponding centroid, the so-called "best matching unit", or BMU. This gives the average distance between each data vector (X) and its BMU and thus measures map resolution:

$$QE = 1/N \sum_{i=1}^{N} \|X_i - (\text{BMU}_{(i)})\| \tag{1}$$

where N is the number of sample vectors x in the image.

This measure completely disregards map topology and alignment, as noted by [8], making it applicable for different kinds and shapes of SOM maps. Besides, the calculation does not rely on any user parameters as seen in (1) above. A 16 by 16 SOM with an initial neighborhood radius of 5 and learning rate of 0.2 was set up for the extraction of data from images. These initial values were arrived at after testing several sizes of the SOM to check that the cluster structures were shown with sufficient resolution and statistical accuracy, [6]. The learning process was started with vectors picked randomly from the image array as the initial values of the model vectors. For each of the following three experiments, the SOM parameters were kept constant.

In this study, we started by applying SOM to time series of original imaging data from a patient's knee before and after blunt force traumatic injury. Then, we added artificial lesion growth to these images and ran SOM analyses on the modified images. [4] modified original images by adding synthetically evolving pathological content of 1%, 5% and 22% volume growth prior to further analyses. They did not use SOM analysis but conducted visual and computational recognition experiments with these images to test the detection of the artificial "pathologies".

## 3. Results from SOM analyses

### 3.1 Original medical images

We used two sets of images from a patient with a sprained knee, courtesy of Hopital de Hautepierre, Strasbourg, France. The same acquisition parameters (machine, sequence, coil, etc) were used to acquire each set which consisted of 20 MRI images. Table 1 shows the QE values obtained from each set of images, taken on two consecutive clinical visits, almost two months apart. Figure 2 is a graphical display of the data.

TABLE 1

FIGURE 2

The QEs shown in Table 1 were submitted to one-way analysis of variance (ANOVA). The difference between image series is statistically significant ($t(1, 38) = 3,336$; $p<.01$).

### 3.2 Medical images with artificially added "lesion" contents

On the first set of images, we added a synthetic lesion to each image to form a second set of images. Since our aim was to discover changes within images between corresponding set of images, the new set of images maintained all the characteristics of the first set, except for the introduced 'lesion' which was uniformly positioned for the 20 images. Thus, the use of synthetic 'lesion' ensures that the differences between sets of images will not be influenced by any external factors like location of camera, lighting, patient position on MRI machine etc. and that the introduced 'growth' is known. The 'lesion' added was a 44 by 26 pixels eclipse-shaped with 72 by 72 dpi resolution, gray-scale and filled with a pattern. A third set of images was similarly created by adding another uniform 'lesion' to the second set. Thus, we create a dataset of images portraying a patient with increasing lesions in his knee. In practice, the three sets of data will have been acquired from the patient on progressive clinical visits. The SOM algorithm was run on each of the three sets of images and the QE was obtained per image as shown in Table 2 and Figure 3.

TABLE 2

FIGURE 3

The results of these simulations show that adding artificial lesion content to the patient's original image data produces a systematic increase in the QE consistent with the increase in lesion contents. The difference in QEs of SOM outputs is statistically significant when SOM analysis of the original image data is compared with SOM analysis of the "double lesion added" image data ($t(1, 38) = 2.055$, $p<.05$). The "single lesion added" treatment, by comparison, did not produce differences in QE that were large enough to reach statistical significance ($t(1,38) = 1.264$, NS). The QEs shown in Table 3 were submitted to one-way analysis of variance (ANOVA). The difference between raw and modified images in a series is statistically significant for both series ($t(1, 38) = 3,337$; $p< .01$ for series 1 and $t(1, 38) = 3,336$; $p< .01$ for series 2). For a graphical representation, see Figure 4.

TABLE 3

FIGURE 4

### 3.3    Medical images with Poisson noise added

We used the Poisson frequency distribution process to add noise on each of the two sets of knee images. Poisson noise was preferred over the other types of impurities generation because it is correlated with the intensity of each pixel in the image. The process produces a sample image from a Poisson distribution for each pixel of the original image. The QE values obtained from each of the original sets and the corresponding noised set are shown in Table 3. The same Poisson distribution parameters were applied to add the impurities in both series. By its nature, Poisson method populates the image with impurities in proportion to existing pixels hence the difference in what was 'added' to each set of images.

## 4. Human detection with random-dot images

*4.1. Objective*

To test whether a systematic increase in the quantization error of the SOM output is, indeed, directly linked to the detectability of potentially critical local image contents, we designed a visual image discrimination experiment using a classic "same-different" paradigm. Images with different percentages of artificially induced and strictly local "lesion" contents (5%, 10 % and 30 %) were paired with original images where no such local "lesion" was added. On each of these images, we ran SOM to determine the quantization error output and to compare its variation with variations in visual change detectability by inexperienced observers. In this experiment, human observers had to judge whether a given image pair was the "same" or "different". Any detection of a difference, called correct positive or "hit", could only be due to detection of the artificially induced local difference ("lesion" content) in one of the images, as all other image parameters (contrast intensity, contrast sign, spatial distribution of contrasts, relative size) were identical in two images of a pair. To determine the subjects' tendency to over-diagnose, we also presented pairs of strictly identical images and recorded the number of false positive detections, or "guesses". The exposure duration of the image pairs was varied to test whether the processing time affects detectability.

*4.2. Subjects*

32 healthy, young male subjects, 26 male and 6 female, all volunteers aged between 19 and 34 years of age participated in this study, which was conducted in conformity with the Helsinki Declaration relative to experiments with human subjects and fully approved by the ethics board of the supervising author's (BDL) host institution (CNRS). All subjects had normal visual acuity and gave written informed consent to participate.

*4.3. Experimental stimuli and procedure*

Computer generated random-dot images of identical size, local contrast (0.7 Michelson contrast) and spatial contrast distribution were created (see Figure 5 for an illustration) using Adobe RGB in Photoshop. In three of these images, one local contrast dot was increased in diameter yielding one image with a 5% local dot size increase, another one with a 10% local dot size increase, and a third one with a 30% local dot size increase, always at exactly the same dot location.

FIGURE 5

Each of these three images was paired with the original "no lesion" image, presented to the left and the right in a pair, in a random order. Images were also paired with their identical images. During the experiment, the subject was seated at a distance of about 75 centimetres from the computer screen in a semi-dark room. The image pairs (see again Figure 2 for an illustration) were presented in a random sequence and each pair was followed by a blank screen presentation of five seconds to avoid visual afterimages, which could have interfered with the task. In one session, the exposure duration for each image pair was five seconds, in another session, the exposure duration was observer controlled. This means that the subject could look at a pair for as long as he deemed necessary to reach a decision, then pressed a key to get the five-second blank screen before the next pair was displayed. The task instruction was to "decide as swiftly and accurately as possible whether two images in a pair appear to be the *same* or *different*. The number of "same" and "different" judgements in response to a given image pair was recorded and written into an individual excel table, for each subject and session. 16 of the 32 subjects started with the five second exposure duration session followed by the session with the observer controlled exposure duration, the other 16 performed the task sessions in the reversed order to counterbalance possible sequential timing effects.

## 5. Results from human detection with random-dot images

### 5.1. Conditional detection rates

The total number of "same" and "different" responses for each type of image pair was divided by the total number of presentations of that pair for a given subject and experimental session. These response frequencies were then multiplied by 100 to produce percentages of correct negatives *CN* reflecting the percentage of "same" responses to pairs of the same image, false negatives *FN* reflecting the percentage of "same" responses to pairs of different images, false positives *FP* ("guesses") reflecting the percentage of "different" responses to pairs of the same image, and correct positives *CP* ("hits") reflecting the percentage of "different" responses to pairs of different images. The distributions are shown in Tables 4, 5, and 6 as a function of the "lesion" contents, with 5%, 10% and 30% local increase in single dot size (1, 2, 3), and as a function of the exposure duration of the image pairs (a and b). We checked to confirm that the position of an image in a pair (*left* or *right*) had no effect on the responses (no positional bias), and average response frequencies for images positioned on *left* and on *right* are shown here.

TABLE 4

TABLE 5

TABLE 6

When comparing between results shown in a) and b) of Tables 4-6, we clearly see that the percentage of false positives *FP*, the so-called "guess rate", does not vary much with the exposure duration of the image pairs, whereas the percentage of correct positives *CP*, the so-called "hit rate", increases markedly when the exposure duration is *ad libitum* and observer controlled. This reveals that the subjects used a constant decision criterion, otherwise the *FP* or "guess rate" would also have varied with the image exposure duration, in the two successive experimental sessions, and that limiting image exposure times negatively affects the CP or "hit rate". When comparing between Tables 4-6, we also quite clearly see that the CP or "hit rate" increases as the "lesion" content in one of the images of a pair increases. In pairs where one of the images has a 5% local dot size ("lesion") increase (Table 4), the "hit rate" *CP* is smaller than the "guess rate" *FP,* which indicates that the subjects are basically guessing and are unable to detect the local difference in image contents. In pairs where one of the images has a 10% or a 30% local dot size ("lesion") increase, the "hit rate" *CP* is twice (Table 5) to three times (Table 6) the "guess rate" *FP,* which shows that the local difference in the image contents is beginning to be detected. In pairs with observer controlled exposure duration where one image has a 30% local increase in "lesion" content, the "hit rate" CP is the highest here at 40%.

*5.2.    Analysis of variance*

In a next step, the average "hit rates" *CP* were submitted to Two-Way ANOVA for the three levels of the "lesion" factor $L_3$ and the two levels of the exposure duration factor $E_2$ to assess the statistical significance of the effects. We observe a statistically significant result for the effect of "lesion" on the average "hit rate", with $F(2, 23) = 38.04$; $p<.001$, and a significant effect of exposure duration, with $F(1, 23) = 8.13$; $p<.05$.

*5. 3.    Comparison with QE values from SOM*

The effect sizes in terms of means and standard errors (SEM) are graphically represented in Figure 6. For comparison of the human detection rates with the QE values from the SOM analyses run on the random-dot images with 5%, 10% and 30% increase in local dot size, we show these QE values as a function of each image type and the correct positive (CP) rates subtracted by the false positive (FP) rates here in Table 7.

TABLE 7

## 6. Discussion

[4] reported that an expert wrongly classified all cases with 1% artificial lesion growth, and only achieved an accuracy of 20% for cases with 5% growth. The same expert, however, correctly classified all cases with a 22% growth. In this study, we introduced a new SOM-based technique for sensing the progression or remission of lesions in medical images. We show that the QE of the SOM output of consecutive analyses of sets of images taken over a time series increases when impurities/lesions on the organ have increased and vice versa. The experiments with human observers confirm that small growths in lesions are hard to detect for humans, while they are reliably captured by the technique introduced in this work. This work is important as it introduces a new technique for the pre-analysis of large bodies of medical images from patients. The technique allows the automatic detection of subtle but significant changes in time series of images likely to reflect growing or receding lesions. In clinical practice, finding evidence for subtle growth through visual inspection of serial imaging can be very difficult. This is especially true for scans taken at relatively short intervals (less than a year). Visual inspection often misses the slow evolution because the change may be obscured by variations in body position, slice position, or intensity profile between scans, as noted by [4]. In some cases, the change can be too small to be noticed, leaving a patient to fate. Surgeons and oncologists frequently compute the change in tumor volume by comparing the measurements of consecutive scans. When the change in tumor volume is too small and hence difficulty to detect between two sequential scans, neuroradiologists tend to compare the most recent scan with the earliest available image to find any visible evidence for the evolution of the tumor. The resulting analysis does not reflect the current development of the tumor but rather a retrospective perspective of the tumor evolution, [4]. Our work takes care of this situation and hence it can aid clinicians in treatment decisions. QE is a quality measure for SOM. It is therefore expected to produce same values when the initial SOM settings and parameters remain the same and there are no changes in the input vector (image). When, the image data is altered and the SOM parameters are not altered, changes in the QE can reasonably be attributed to the

developments taking place in the organ whose image is under study. This is why we have proposed QE as a clinical determinant of the progression or remission of lesions in medical images. We hope to carry on with further experiments in this area especially with real images, comparing between results from real patient data. We also expect to confirm our simulation results in the light of analysis by human experts and metrics proposed by the World Health Organization [9].

## 7. Conclusions

When the QE of a patient's images taken at different consecutive times rises, it is a potential indication that lesions or impurities on the organ under study are increasing, while a decrease may indicate the lesions are receding. A common approach to measuring many cellular processes by image analysis is to start with segmenting the image into components of interest. We have suggested a method to follow-up a patient treatment after diagnosis that does not rely on data from the segments only, but performs a global analysis of the entire image. Monitoring cancer progression/remission is often estimated via manual segmentation of several images in an MRI sequence, which is prohibitively time consuming, or via automatic segmentation, which is a challenging and computationally expensive task that may result in high estimation errors [10]. In this study, we have estimated the disease progression in real-time directly from image statistics using a self-organizing machine learning technique. We demonstrate that the QE value of the output of these analyses "detects" the smallest increase in potentially relevant local image contents that are impossible for humans to see.

**Funding**

This research did not receive any specific grant from funding agencies in the public, commercial, or not-for-profit sectors.

**Acknowledgments**

We thank our colleague, Dr Philippe Choquet from Hôpital de Hautepierre, and ICube UMR 7357 CNRS-UdS, Strasbourg France, who provided valuable insight and expertise for this research.

**Figures and Tables with legends**

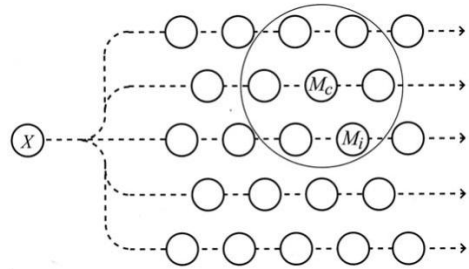

Figure 1: Schematic illustration of a self-organizing map. An input data item X is broadcast to a set of models $M_i$, of which $M_c$ matches best with X. All models that lie in the neighborhood (larger circle) of $M_c$ in the grid match better with X than with the rest, from [6].

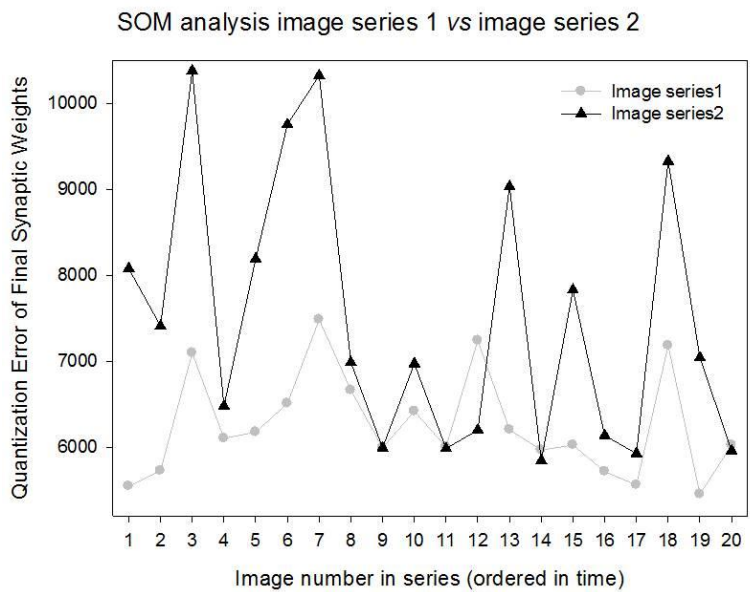

Figure 2: Results from a series of SOM analyses on time series of knee images, taken at two different moments in time. It is shown that the QE in the SOM output increases significantly (t (1, 38) = 3,336; p<.01) between image series taken before (series 1) and after (series 2).

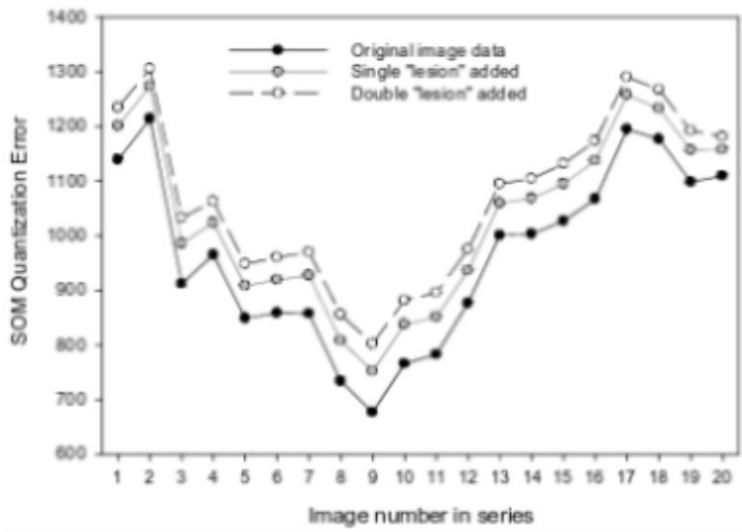

Figure 3: Graphical comparison of QE values of images with increasing lesions.

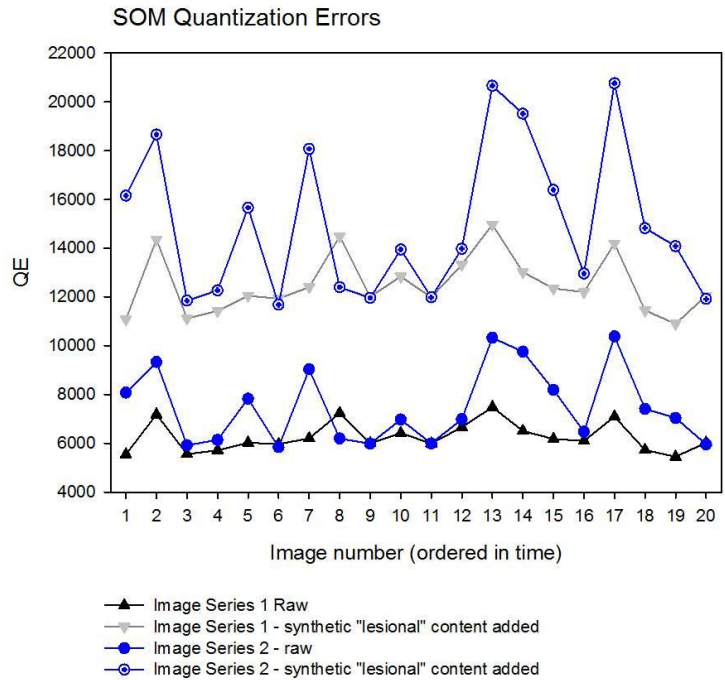

Figure 4: Results from a series of SOM analyses on time series of images, taken at two different moments in time. It is shown that the QE in the SOM output increases significantly (t (1, 38) = 3,336; p<.01) between the image series. In the raw images series 1, small synthetic lesion was added, while in series 2, a larger synthetic lesion was added. For each manipulation, the difference in the QE from the SOM outputs is statistically significant (t (1, 38) = 3,337; p< .01 for series 1 and t (1, 38) = 3,336 ; p< .01 for series 2).

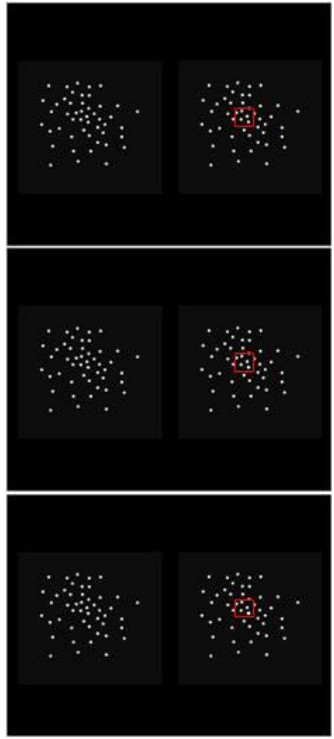

Figure 5: Three random dot-images with different percentages of artificially induced and strictly local "lesion" contents (5%, 10 % and 30 % increase in size of a single small dot, shown here highlighted by the red square) were paired with original image where no such local "lesion" was added (images on left in a given pair here above). Right and left images in a pair varied between presentations, in random order. Pairs of identical images (not shown here) were also presented in a task sequence to measure false alert rates ("guess rates").

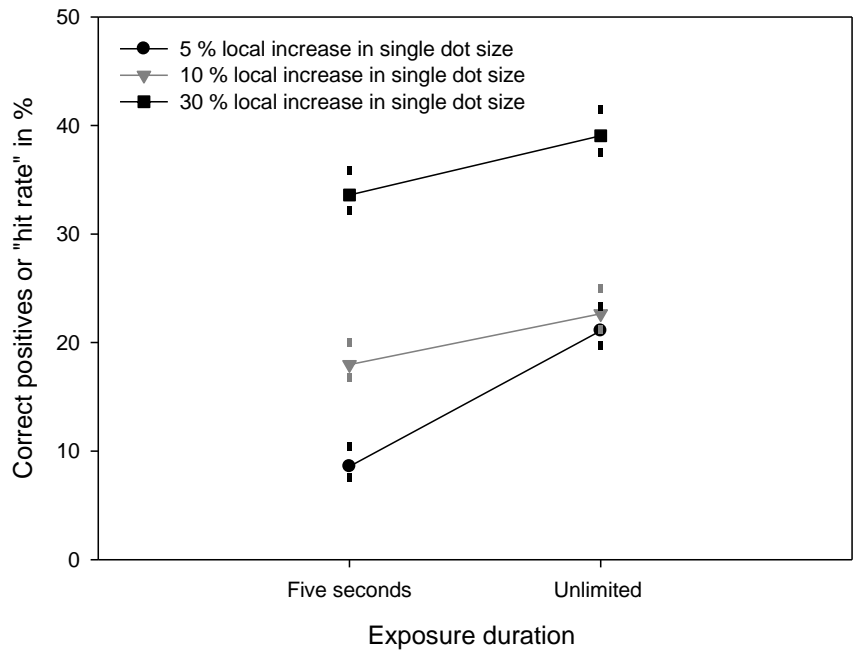

Figure 6: Average "hit rates" (*CP*) and standard errors as a function of the "lesion" content and the exposure duration of image pairs.

| Image | QE 1st | QE 2nd |
|---|---|---|
| dcm 0001 | 5544.68 | 8078.32 |
| dcm 0002 | 5724.76 | 7410.38 |
| dcm 0003 | 7096.77 | 10381.9 |
| dcm 0004 | 6101.77 | 6478.89 |
| dcm 0005 | 6174.82 | 8193.23 |
| dcm 0006 | 6507.84 | 9757.81 |
| dcm 0007 | 7484.48 | 10326.94 |
| dcm 0008 | 6661.52 | 6985.06 |
| dcm 0009 | 5992.41 | 5992.17 |
| dcm 0010 | 6417.38 | 6972.39 |
| dcm 0011 | 6001.4 | 5982.37 |
| dcm 0012 | 7240.49 | 6198.58 |
| dcm 0013 | 6201.82 | 9034.32 |
| dcm 0014 | 5966.33 | 5842.39 |
| dcm 0015 | 6024.03 | 7830.31 |
| dcm 0016 | 5714.79 | 6135.71 |
| dcm 0017 | 5557.94 | 5924.59 |
| dcm 0018 | 7182.26 | 9330.04 |
| dcm 0019 | 5450.78 | 7041.98 |
| dcm 0020 | 6023.86 | 5957.58 |

Table 1: QE values from images taken from two consecutive clinical visits by a patient with an injured left-leg knee. There is an increase in QE values between each image in the two series.

| Image | Original | 1 "lesion" | 2 "lesions" |
|---|---|---|---|
| dcm 0001 | 1138.9128 | 1200.9820 | 1234.8677 |
| dcm 0002 | 1213.9390 | 1273.5073 | 1305.3644 |
| dcm 0003 | 912.0454 | 985.4192 | 1032.4355 |
| dcm 0004 | 965.0731 | 1024.0330 | 1062.7660 |
| dcm 0005 | 848.7616 | 908.4071 | 948.0895 |
| dcm 0006 | 858.5535 | 919.0936 | 960.0879 |
| dcm 0007 | 857.2325 | 927.1354 | 969.5507 |
| dcm 0008 | 734.0570 | 808.2034 | 855.7769 |
| dcm 0009 | 676.9681 | 751.9430 | 802.0007 |
| dcm 0010 | 765.6439 | 837.8734 | 881.9957 |
| dcm 0011 | 782.6192 | 851.3009 | 895.5168 |
| dcm 0012 | 876.5664 | 935.8310 | 974.4636 |
| dcm 0013 | 1000.3647 | 1059.5208 | 1095.0401 |
| dcm 0014 | 1003.1925 | 1068.2832 | 1104.3974 |
| dcm 0015 | 1026.7828 | 1095.1051 | 1131.4206 |
| dcm 0016 | 1067.1361 | 1137.2907 | 1172.9960 |
| dcm 0017 | 1194.5449 | 1257.6472 | 1290.4847 |
| dcm 0018 | 1176.3578 | 1232.5629 | 1267.2867 |
| dcm 0019 | 1098.3993 | 1156.7749 | 1191.4239 |
| dcm 0020 | 1109.3291 | 1157.2493 | 1181.3063 |

Table 2: QE distributions for original patient images and images with synthetic lesions added. QE values increase with increase of 'lesions' added to an image.

| 1st clinical | Dots added | 2nd clinical | Dots added |
|---|---|---|---|
| 5544.4807 | 11086.7877 | 8078.2439 | 16157.1898 |
| 7181.9884 | 14364.0413 | 9330.1503 | 18660.5707 |
| 5558.1511 | 11117.9896 | 5924.6644 | 11850.9655 |
| 5714.7921 | 11429.2792 | 6135.6891 | 12273.3048 |
| 6023.7532 | 12048.4203 | 7830.3322 | 15663.1586 |
| 5966.3444 | 11932.9385 | 5842.4854 | 11684.4111 |
| 6201.7292 | 12405.2023 | 9034.2843 | 18067.9178 |
| 7240.853 | 14482.1577 | 6198.6079 | 12401.7001 |
| 6001.4699 | 12002.7776 | 5982.453 | 11965.1671 |
| 6417.1673 | 12836.4429 | 6972.2216 | 13943.6979 |
| 5992.5 | 11984.7077 | 5992.1492 | 11984.4634 |
| 6661.4586 | 13324.6943 | 6985.1401 | 13973.7904 |
| 7484.6984 | 14968.1884 | 10327.1069 | 20659.262 |
| 6507.8144 | 13017.8913 | 9757.7571 | 19514.2742 |
| 6174.883 | 12349.7042 | 8193.1116 | 16388.2526 |
| 6101.946 | 12203.2147 | 6478.8401 | 12960.4942 |
| 7096.3922 | 14191.5997 | 10381.9172 | 20764.6702 |
| 5724.8007 | 11450.6902 | 7410.2646 | 14823.074 |
| 5450.6741 | 10901.116 | 7041.9858 | 14083.6771 |
| 6023.8499 | 12049.4355 | 5957.4332 | 11916.3526 |

Table 3: The QE values of the set of images taken from the patient on the first clinical visit, 29th April 2016, in the 1st column and in the 2nd column are the QE values of the same images with added dots. The QE values in the 3rd column are from images taken from the patient on the second clinical visit, 17th June 2016. The 4th column shows the QE values of the resulting images after adding dots. The dots were added to each image based on Poisson distribution frequency.

*Image pairs with five seconds exposure*

|  |  | SAME | DIFFERENT |
|---|---|---|---|
| **R** | "same" | **88.7** *(CN)* | **91.4** *(FN)* |
|  | "different" | **11.3** *(FP)* | **8.6** *(CP)* |

a)

*Image pairs with observer controlled exposure*

|  |  | SAME | DIFFERENT |
|---|---|---|---|
| **R** | "same" | **86.5** *(CN)* | **91.4** *(FN)* |
|  | "different" | **13.5** *(FP)* | **8.6** *(CP)* |

b)

<u>Table 4:</u> Conditional response rates *R* in percent (%) for "no-lesion" images paired with "5% lesion" images under conditions of five seconds exposure duration *(a)*, and observer controlled exposure duration *(b)* for each image pair. Correct positive (*CP*), often also called "hits", correct negative (*CN*), false positive (*FP*), and false negative (*FN*) response rates are shown.

*Image pairs with five seconds exposure*

|   |           | SAME        | DIFFERENT   |
|---|-----------|-------------|-------------|
| **R** | "same"    | **87.5** *(CN)* | **82.0** *(FN)* |
| **R** | "different" | **12.5** *(FP)* | **18.0** *(CP)* |

a)

*Image pairs with observer controlled exposure*

|   |           | SAME        | DIFFERENT   |
|---|-----------|-------------|-------------|
| **R** | "same"    | **87.0** *(CN)* | **77.4** *(FN)* |
| **R** | "different" | **13.0** *(FP)* | **22.6** *(CP)* |

b)

<u>Table 5:</u> Conditional response rates (%) for "no-lesion" images paired with "10% lesion" images under conditions of five seconds exposure duration *(a)*, and observer controlled exposure duration *(b)* for each image pair. Correct positive (*CP*), often also called "hits", correct negative (*CN*), false positive (*FP*), and false negative (*FN*) response rates are shown.

*Image pairs with five seconds exposure*

|   |           | SAME        | DIFFERENT   |
|---|-----------|-------------|-------------|
|   | "same"    | **85.5** *(CN)* | **66.4** *(FN)* |
| *R* | "different" | **14.5** *(FP)* | **33.6** *(CP)* |

a)

*Image pairs with observer controlled exposure*

|   |           | SAME        | DIFFERENT   |
|---|-----------|-------------|-------------|
|   | "same"    | **86.5** *(CN)* | **60.9** *(FN)* |
| *R* | "different" | **13.5** *(FP)* | **39.1** *(CP)* |

b)

Table 6: Conditional response rates (%) for "no-lesion" images paired with **"30% lesion"** images under conditions of five seconds exposure duration *(a)*, and observer controlled exposure duration *(b)* for each image pair. Correct positive (*CP*), often also called "hits", correct negative (*CN*), false positive (*FP*), and false negative (*FN*) response rates are shown.

| Lesion % increase in local dot size | QE value generated by SOM output | *Detection* for five second exposure duration | *Detection* for observer controlled duration |
|---|---|---|---|
| 0% (original) | 750.3749 | | |
| 5% | 750.4555 | 8.6 % - 13 % | 8.6 % - 13 % |
| 10% | 751.7827 | 18 % - 13 % | 22.6 % - 13 % |
| 30% | 754.4679 | 33.6 % - 13 % | 39.1 % - 13 % |

Table 7: QE values and average detection rates for lesion detection by human observers as a function of the lesion content in % local dot size increase in one of the images in a pair and exposure duration. The average rate of false positives *("guess rate")* was 13% in response to image pairs containing two identical images with no lesion content. To conclude about detection, the average rate of false positives is to be subtracted, as here above, from the average rate of correct positives, as stated in [11].